\documentclass[10pt,twocolumn,letterpaper]{article}

\usepackage{cvpr}      
\definecolor{cvprblue}{rgb}{0.21,0.49,0.74}
\usepackage[pagebackref=true,breaklinks=true,colorlinks=true,bookmarks=false]{hyperref}
\definecolor{DirtyOrange}{HTML}{D87C46}
\hypersetup{linkcolor=DirtyOrange}
\hypersetup{urlcolor=DirtyOrange}
\definecolor{Indigo}{HTML}{4B0082}
\hypersetup{citecolor=Indigo}
\usepackage{algorithm}  
\usepackage{algpseudocode}
\usepackage{amsmath} 
\usepackage{float}
\usepackage{amsthm}

\usepackage{graphicx}
\usepackage{xcolor}
\usepackage{colortbl}
\usepackage{multirow}
\definecolor{green1}{HTML}{b4dbca}
\definecolor{green2}{HTML}{daede4}
\definecolor{deepgreen}{HTML}{235d3a}
\definecolor{red1}{HTML}{ffa8a8}
\definecolor{red2}{HTML}{ffe3e3}
\definecolor{gray2}{HTML}{eeecea}
\definecolor{purple1}{HTML}{E2C4FD}
\definecolor{purple2}{HTML}{EEE6FD}
\definecolor{pink}{HTML}{FFCCB0}
\usepackage{booktabs}

\def\paperID{} 
\def\confName{CVPR}
\def\confYear{2026}

\title{NTK-Guided Implicit Neural Teaching}

\author{Chen Zhang\thanks{Equal contribution}\,\,~\thanks{Corresponding author} \quad Wei Zuo\footnote[1]{} \quad Bingyang Cheng \quad  Yikun Wang \quad Wei-Bin Kou \\
\vspace{0.2cm}
Yik-Chung Wu \quad Ngai Wong  \\
\vspace{0.2cm}
The University of Hong Kong \\
\texttt{\small \{\href{mailto:czhang6@connect.hku.hk}{\color{black}czhang6}, \href{mailto:weizuo002@connect.hku.hk}{\color{black}weizuo002}\}@connect.hku.hk}\quad\texttt{\small \{\href{mailto:ycwu@eee.hku.hk}{\color{black}ycwu}, \href{mailto:nwong@eee.hku.hk}{\color{black}nwong}\}@eee.hku.hk}
\\[8pt]
  \centerline{%
    \raisebox{-0.2\height}{\includegraphics[height=1.3em]{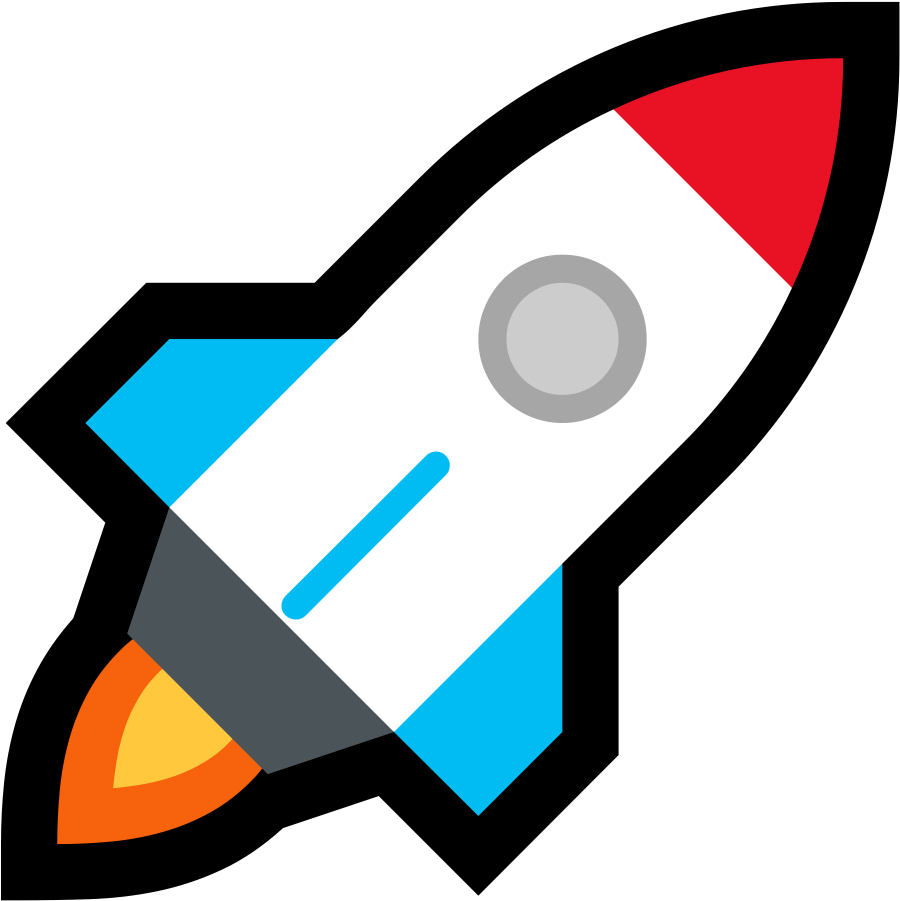}}\ \href{https://chen2hang.github.io/_publications/ntk_guided_implicit_neural_teaching/nint.html}{\texttt{\large Project page}} 
  }
}

\begin{document}
\maketitle
\begin{abstract}
Implicit Neural Representations (INRs) parameterize continuous signals via multilayer perceptrons (MLPs), enabling compact, resolution-independent modeling for tasks like image, audio, and 3D reconstruction. However, fitting high-resolution signals demands optimizing over millions of coordinates, incurring prohibitive computational costs. To address it, we propose NTK-Guided Implicit Neural Teaching (NINT), which accelerates training by dynamically selecting coordinates that maximize global functional updates. Leveraging the Neural Tangent Kernel (NTK), NINT scores examples by the norm of their NTK-augmented loss gradients, capturing both fitting errors and heterogeneous leverage (self-influence and cross-coordinate coupling). This dual consideration enables faster convergence compared to existing methods. Through extensive experiments, we demonstrate that NINT significantly reduces training time by nearly half while maintaining or improving representation quality, establishing state-of-the-art acceleration among recent sampling-based strategies.
\end{abstract}
    
\section{Introduction}
\label{sec:intro}

Implicit Neural Representations (INRs)~\cite{sitzmann2020implicit,tancik2020fourier} model discrete signals (\eg, audios, images, or 3D scenes) using continuous multilayer perceptrons (MLPs). These MLPs take low-dimensional coordinates as input (\eg, pixel locations or spatiotemporal points) and output the corresponding signal values, effectively learning a continuous function that interpolates the discrete data with high fidelity. This coordinate-based formulation has revolutionized several domains, including high-resolution image representation~\cite{sitzmann2020implicit,reddy2021multi}, novel view synthesis~\cite{mildenhall2021nerf,martin2021nerf}, and compact signal compression~\cite{dupont2021coin,pistilli2022signal,strumpler2022implicit,schwarz2023modality}.

Nevertheless, training INRs is computationally intensive due to the large number of training examples inherent in high-resolution or high-dimensional signals. For instance, a single $1024\times1024$ image contains over one million pixels, each treated as an independent training point. Videos and 3D volumes scale this further into billions of coordinates. As a result, standard gradient descent requires repeated full passes over the entire dataset, making training slow and resource-heavy.

\begin{figure}[t]
    \centering
    \includegraphics[width=0.48\textwidth]{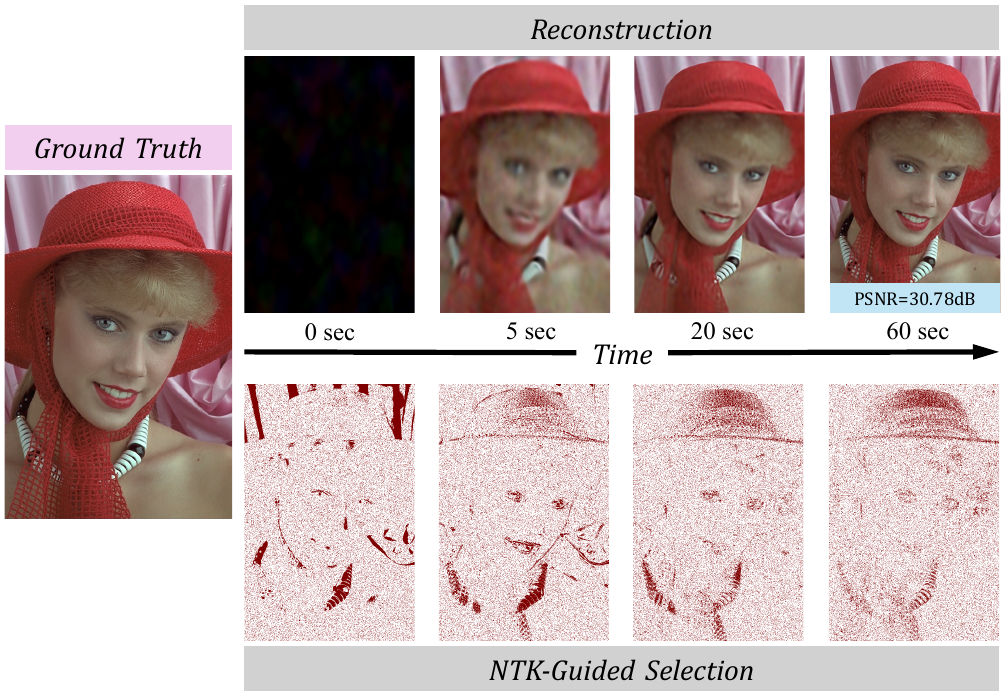}
    \caption{NINT selection and reconstruction on \textit{kodim04} from Kodak~\cite{kodak}. \textbf{Top}: INR reconstructions at 0, 5, 20, 60 sec (final PSNR 30.78 dB). \textbf{Bottom}: NINT-selected coordinates (red), scored by NTK-augmented loss gradient norm to capture fitting errors and heterogeneous leverage (self-influence and cross-coupling).}
    \label{fig:process}	\vskip -0.2in
\end{figure}

Several acceleration strategies have been proposed, each with inherent limitations. Partition-based methods~\cite{martel2021acorn,saragadam2022miner,liu2023partition} divide the signal domain across multiple specialized MLPs, but this increases architectural complexity and inference overhead. Hybrid explicit-implicit approaches~\cite{muller2022instant,chen2022tensorf,xie2023diner} augment MLPs with structured representations such as tensors or voxels, achieving faster convergence at the expense of higher memory consumption. Meta-learning techniques~\cite{tancik2021learned,chen2022transformers} accelerate fitting via pre-trained initialization, yet demand large homogeneous datasets for pre-training, limiting flexibility and requiring significant upfront computation.

In contrast, sampling-based methods~\cite{gai2023egra,zhang2024nonparametric,zhang2025expansive,zhang2025evos} offer a lightweight alternative by selectively training on a subset of coordinates each iteration. These approaches reduce per-step computation without modifying the model or requiring auxiliary data. However, most rely on static heuristics, such as the magnitude of current output error or local signal variation, to guide selection. While intuitive, these criteria overlook the evolving dynamics of the MLP during training, particularly how different coordinates influence parameter updates and global convergence. This limitation restricts their ability to achieve maximal acceleration.

To address these limitations, we propose NTK-Guided Implicit Neural Teaching (NINT), a novel sampling framework that incorporates the Neural Tangent Kernel (NTK) to capture the MLP’s evolving training dynamics. Rather than relying solely on static measures of output error, NINT jointly evaluates two complementary factors: (1) the gradient of the loss with respect to the network output, which identifies regions of high fitting discrepancy, and (2) the NTK-induced influence of each coordinate on parameter updates, which quantifies how strongly a point drives global model changes during training. By prioritizing examples that combine high fitting error with strong dynamic influence, NINT ensures that each training batch maximally accelerates global convergence. Fig.~\ref{fig:process} visualizes the evolving reconstructions and NINT-selected coordinates. Through extensive experiments, NINT reduces training time by nearly half compared to full-dataset training, while preserving or improving final representation quality.
Among recent sampling-based strategies~\cite{gai2023egra,zhang2025expansive,zhang2025evos,zhang2024nonparametric,kheradmand2024accelerating}, NINT establishes new state-of-the-art performance in both speed and fidelity. Our key contributions are:
\begin{itemize}[leftmargin=*,nosep]
	\setlength\itemsep{0.56em}
    \item \textbf{NTK-centric analysis of INR dynamics}: We derive the functional evolution under gradient descent, exposing the flaws in error-only sampling due to neglected self-leverage and cross-coordinate coupling (Sec.~\ref{sec:ntk}).
    \item \textbf{NINT sampling strategy}: A principled, plug-and-play method that selects examples by maximizing the magnitude of the functional update induced by each coordinate, computed as the norm of NTK row (capturing global influence) multiplied by the loss gradients, ensuring every parameter step drives the largest possible improvement across the entire signal (Sec.~\ref{sec:nint}).
    \item \textbf{State-of-the-art acceleration}: Extensive experiments demonstrate that NINT reduces training time by nealy half compared to full-batch baselines, while matching or exceeding reconstruction quality, outperforming recent sampling-based strategies (Sec.~\ref{sec:results}).
\end{itemize}

\section{Related Works}
\label{sec:related}

\subsection{Implicit Neural Representations}
Implicit Neural Representations (INRs)~\cite{sitzmann2020implicit,park2019deepsdf,atzmon2020sal,gropp2020implicit} encode discrete signals as continuous functions using coordinate-based MLPs, delivering memory-efficient, high-resolution representations across diverse domains, spanning images~\cite{dupont2021coin,xie2023diner}, videos~\cite{chen2021nerv,kim2022scalable,maiya2023nirvana}, 3D shapes~\cite{chabra2020deep,li2023neuralangelo}, scenes~\cite{mildenhall2020nerf,muller2022instant,chen2022tensorf}, and unconventional data types~\cite{huang2023compressing,bandyopadhyay2024sketchinr}. Beyond reconstruction, INRs drive breakthroughs in compression~\cite{strumpler2022implicit}, generative modeling~\cite{park2024ddmi}, novel view synthesis~\cite{mildenhall2020nerf}, inverse imaging~\cite{huang2023inverting}, copyright protection~\cite{luo2023copyrnerf,song2024protecting,luo2025nerf,luo2025imagesentinel}, and physics-informed PDE solving~\cite{raissi2019physics}. Progress in architectural design, including sinusoidal activations~\cite{saragadam2023wire,ramasinghe2022beyond}, Fourier positional encodings~\cite{tancik2020fourier,liu2023finer}, and structured priors~\cite{yuce2022structured,wang2022neural,grattarola2022generalised,lindell2022bacon,li2023scone,molaei2023implicit,2024asmr}, has significantly boosted expressiveness and alleviated spectral bias~\cite{rahaman2019spectral}. Yet, training on complex signals remains computationally demanding due to massive coordinate counts, highlighting the critical need for efficient, model-agnostic acceleration methods.

\subsection{Acceleration for INR Training}

To mitigate the high computational cost of training over vast coordinate sets, prior works have explored diverse acceleration paradigms while trading off model complexity, memory, or auxiliary resources. Partition-based techniques~\cite{liu2023partition} segment the domain into subregions managed by specialized MLPs, leveraging regular grids~\cite{reiser2021kilonerf}, adaptive partitioning~\cite{martel2021acorn}, Voronoi diagrams~\cite{rebain2021derf}, or pyramidal structures~\cite{saragadam2022miner}, yet they introduce inference latency from ensemble coordination. Explicit caching methods integrate structured priors, such as hash grids~\cite{muller2022instant}, low-rank tensors~\cite{chen2022tensorf}, tree hierarchies~\cite{liu2020neural,yang2023tinc}, or point clouds~\cite{xu2022point}, to expedite convergence at the cost of increased memory. Meta-learning approaches~\cite{tancik2021learned,chen2022transformers} provide task-specific initializations through pre-training on vast corpora~\cite{tancik2021learned} or hypernetworks~\cite{sitzmann2020implicit,szatkowski2023hypernetworks}, but demand substantial upfront data and computation. Additional efficiency gains have been pursued via modulators~\cite{mehta2021modulated}, reparameterizations~\cite{shi2024improved}, input transformations~\cite{seo2024search}, and normalization layers~\cite{cai2024batch}.

In contrast, sampling-based methods~\cite{gai2023egra,zhang2024nonparametric,zhang2025expansive,zhang2025evos} preserve architectural simplicity by subsampling coordinates per iteration, including edge-aware EGRA~\cite{gai2023egra}, frequency-prior-driven Expansive Supervision~\cite{zhang2025expansive}, Monte-Carlo Soft Mining~\cite{kheradmand2024accelerating}, and evolutionary EVOS~\cite{zhang2025evos}. Most closely related, INT~\cite{zhang2024nonparametric} recasts INR acceleration as a nonparametric teaching problem (\ie, strategically selecting the most informative coordinates to guide MLP convergence, akin to a teacher curating high-impact examples)~\cite{zhang2023nonparametric,zhang2023mint,zhang2025nonparametric,zhang2026nonparametric}, using greedy functional algorithms for coordinate selection. While effective, these heuristics typically ignore the MLP's parameter update dynamics, constraining convergence speed. As discussed in \S\ref{sec:ntk}, INT is essentially using a matrix similar to an identity matrix to approximate the NTK, with nearly identical values along the diagonal. This inaccurately models the true direction of parameter updates unless the NTK is nearly diagonal, which is rarely the case in typical MLP training.~\cite{jacot2018neural,lee2019wide}. Consequently, the selected coordinates may not optimally drive convergence. Differently, our NINT integrates the full NTK to explicitly model training dynamics in the selection process, jointly optimizing for high output-gradient error and strong parameter-update influence, thereby maximizing per-batch progress toward global convergence.

\section{Method}
\label{sec:method}

We begin by formulating the implicit neural representation (INR) as an MLP that maps coordinates to signal attributes, trained via empirical risk minimization. We then analyze INR training dynamics through the lens of the neural tangent kernel (NTK), revealing heterogeneous self-leverage and functional coupling that prior error-only sampling strategies fail to account for. To address this, we propose NTK-Guided Implicit Neural Teaching (NINT), which selects influential coordinates by maximizing NTK-augmented gradients for faster global convergence. We detail the formulation in Sec.~\ref{sec:formulation}, dynamics in Sec.~\ref{sec:ntk}, and NINT in Sec.~\ref{sec:nint}.

\subsection{Formulation}
\label{sec:formulation}

Let $S = \{(\mathbf{x}_i, \mathbf{y}_i)\}_{i=1}^N$ denote a natural signal consisting of $N$ observation pairs, where $\mathbf{x} \in \mathbb{R}^{m}$ represents an $m$-dimensional coordinate (\eg, spatial, temporal, or spatiotemporal location) and $\mathbf{y}_i \in \mathbb{R}^n$ denotes the corresponding $n$-dimensional attribute vector (\eg, audio intensity, pixel values, density, or feature embedding).

The goal of INR is to learn a continuous mapping $f_\theta: \mathbb{R}^m \to \mathbb{R}^n$ parameterized by an MLP with parameters $\theta$, such that $f_\theta(\mathbf{x}_i) \approx \mathbf{y}_i$ for all $i \in \{1, \dots, N\}$, and ideally, $f_\theta$ generalizes to unobserved coordinates $\mathbf{x} \notin \{\mathbf{x}_i\}_{i=1}^N$. This is achieved by minimizing an empirical risk over the observed data:
\begin{eqnarray} \label{eq:opti}
    \theta^\star = \arg\min_{\theta} \frac{1}{N} \sum_{i=1}^N \mathcal{L}\left(f_\theta(\mathbf{x}_i), \mathbf{y}_i\right),
\end{eqnarray}
where $\mathcal{L}: \mathbb{R}^n \times \mathbb{R}^n \to \mathbb{R}_+$ is a task-specific loss function (typically the $\ell_2$ norm for regression). A regularizer $\mathcal{R}(\theta)$ with coefficient $\lambda \geq 0$ may be included to enhance generalization \cite{li2023regularize}.

The MLP $f_\theta$ typically consists of $L$ fully connected layers with non-linear activations (\eg, sine \cite{sitzmann2020implicit}):
\begin{eqnarray}
    &&\mathbf{h}_0 = \mathbf{x}\nonumber\\
    &&\mathbf{h}_\ell = \sigma_\ell(W_\ell \mathbf{h}_{\ell-1} + \mathbf{b}_\ell), \quad \ell = 1,\dots,L-1,\nonumber\\
    &&f_\theta(\mathbf{x}) = W_L \mathbf{h}_{L-1} + \mathbf{b}_L
\end{eqnarray}
where $\{W_\ell, \mathbf{b}_\ell\}\equiv\theta$ are learnable weights and biases, and $\sigma_\ell$ denotes the activation function at layer $\ell$. To enhance high-frequency reconstruction, positional encoding may be applied to the input~\cite{mildenhall2020nerf,tancik2020fourier,shi2024improved}.

During inference, $f_\theta$ enables continuous evaluation at arbitrary resolutions without storing the full signal, making INRs particularly suitable for high-dimensional, large-scale, or irregularly sampled data. The learned representation is \textit{implicit}, as the signal is encoded entirely within the network parameters $\theta$, rather than in discrete voxels or vertices.

\subsection{INR Training Dynamics}

Training an INR as defined in Eq.~\ref{eq:opti} is typically performed via gradient descent on the network parameters~\cite{ruder2016overview}. For a dataset of size $N$ (\ie, $N$ observation pairs), the full-batch parameter update at iteration $t$ is:
\begin{equation}
\theta^{t+1} \gets \theta^t - \frac{\eta}{N} \sum_{i=1}^N \nabla_{\theta} \mathcal{L}\bigl(f_{\theta^t}(\mathbf{x}_i), \mathbf{y}_i\bigr),
\end{equation}
where $\eta > 0$ is the learning rate.

When $\eta$ is sufficiently small, the discrete update can be approximated in continuous time as a differential equation:
\begin{equation} \label{eq:paraevo}
\frac{\partial \theta^t}{\partial t} = -\frac{\eta}{N} {\big[\nabla_\theta f_{\theta^t}(\mathbf{x}_i)\big]_{i=1}^N}^\top \cdot \big[ \nabla_f \mathcal{L}\bigl(f_{\theta^t}(\mathbf{x}_i), \mathbf{y}_i\bigr) \big]_{i=1}^N.
\end{equation}

To understand how the \emph{function} $f_{\theta^t}$ evolves, consider its first-order Taylor expansion:
\begin{equation}
f_{\theta^{t+1}}(\mathbf{x}) - f_{\theta^t}(\mathbf{x}) = \Big\langle \frac{\partial f_{\theta^t}(\mathbf{x})}{\partial \theta^t}, \theta^{t+1} - \theta^t \Big\rangle + o(\|\theta^{t+1} - \theta^t\|).
\end{equation}
Taking the continuous-time limit yields:
\begin{equation}
\frac{\partial f_{\theta^t}(\mathbf{x})}{\partial t} \simeq \Big\langle \frac{\partial f_{\theta^t}(\mathbf{x})}{\partial \theta^t}, \frac{\partial \theta^t}{\partial t} \Big\rangle.
\end{equation}

Substituting the parameter evolution (\ie, Eq. \ref{eq:paraevo}) gives:
\begin{equation}    
    \label{eq:func_evol}
\frac{\partial f_{\theta^t}(\mathbf{x})}{\partial t} \simeq -\frac{\eta}{N} {\big[ K_{\theta^t}(\mathbf{x}_i, \mathbf{x})\big]_{i=1}^N}^\top \cdot \big[ \nabla_f \mathcal{L}\bigl(f_{\theta^t}(\mathbf{x}_i), \mathbf{y}_i\bigr) \big]_{i=1}^N,
\end{equation}
where
\begin{equation}\label{eq:ntk}
K_{\theta^t}(\mathbf{x}_i, \mathbf{x}) \equiv \Big\langle \frac{\partial f_{\theta^t}(\mathbf{x}_i)}{\partial \theta^t}, \frac{\partial f_{\theta^t}(\mathbf{x})}{\partial \theta^t} \Big\rangle
\end{equation}
is the Neural Tangent Kernel (NTK) at parameters $\theta^t$, input $\mathbf{x}_i$ and $\mathbf{x}$~\cite{jacot2018neural}.

In the infinite-width limit, $K_{\theta^t}$ remains \emph{constant} throughout training, transforming gradient descent into kernel regression in function space~\cite{jacot2018neural}. For finite-width MLPs, the NTK evolves gradually but remains a reliable descriptor of training dynamics~\cite{lee2019wide,arora2019exact}.

The full NTK matrix $K_{\theta^t}(\mathbf{x}_i, \mathbf{x}_j)$ captures \textbf{functional coupling} between coordinates via inner products in parameter-gradient space: the off-diagonal entry $K_{\theta^t}(\mathbf{x}_i, \mathbf{x}_j)$ measures the projection of $\frac{\partial f_{\theta^t}(\mathbf{x}_j)}{\partial \theta^t}$ onto the update direction induced by $\mathbf{x}_i$. A large $|K_{\theta^t}(\mathbf{x}_i, \mathbf{x}_j)|$ indicates that a parameter update driven by the loss at $\mathbf{x}_i$ will significantly \textbf{co-move} the output at $\mathbf{x}_j$, even if $\mathbf{x}_j$ is not included in the current batch.

Meanwhile, the diagonal trace $K_{\theta^t}(\mathbf{x}_i, \mathbf{x}_i) = \big\| \frac{\partial f_{\theta^t}(\mathbf{x}_i)}{\partial \theta^t} \big\|_2^2$ quantifies the \textbf{self-leverage} of $\mathbf{x}_i$, \ie, the squared norm of its output gradient in parameter space. High $K_{\theta^t}(\mathbf{x}_i, \mathbf{x}_i)$ implies that a unit parameter update along the loss gradient at $\mathbf{x}_i$ induces a large change in its own output $f_\theta$ and, through off-diagonal coupling, across the global function.

\begin{figure}[t]
    \centering
    \includegraphics[width=0.48\textwidth]{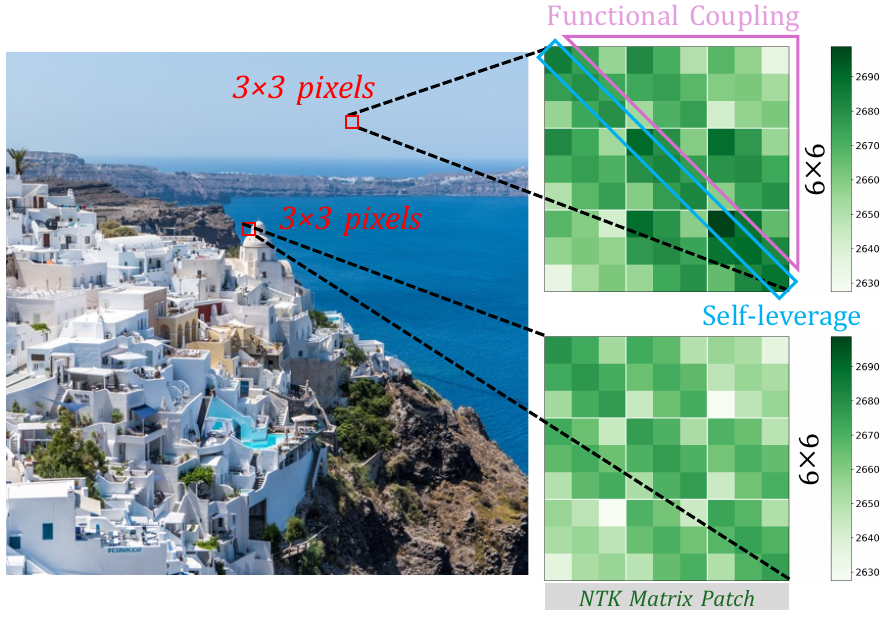}
    \caption{NTK visualization on the image 02 from DIV2K~\cite{agustsson2017ntire}. Two $9 \times 9$ NTK patches correspond to two $3 \times 3$ pixel regions (red). Strong off-diagonals show significant functional coupling between regions, while diverse diagonals indicate heterogeneous self-leverage. Color denotes magnitude.}
    \label{fig:ntksel}
\end{figure}

We illustrate this NTK behavior in Fig.~\ref{fig:ntksel}, where high self-leverage (diagonal) and strong functional coupling (off-diagonal) are evident in local $9 \times 9$ NTK matrix patches centered on selected $3 \times 3$ pixel regions.

This dual structure, \ie, \textbf{varying self-leverage} and \textbf{non-negligible functional coupling}, exposes a critical flaw in prior sampling methods~\cite{gai2023egra,zhang2024nonparametric,zhang2025evos}. By selecting coordinates solely based on output error $\|\nabla_f \mathcal{L}\|_2$, these approaches implicitly assume the NTK is \textbf{diagonal and isotropic}: \ie, no cross-coordinate influence ($K_{\theta^t}(\mathbf{x}_i, \mathbf{x}_j) \approx 0$ for $i \neq j$) and uniform self-leverage ($K_{\theta^t}(\mathbf{x}_i, \mathbf{x}_i) \approx c$ for all $i$). This is equivalent to approximating the NTK with a scaled identity matrix. 

\label{sec:ntk}

In practice, neither holds: MLPs exhibit strong off-diagonal coupling due to weight sharing~\cite{jacot2018neural}, and diagonal values vary by orders of magnitude—coordinates in high-frequency or high-curvature regions (\eg, edges, transients) have significantly larger NTK traces than those in smooth or noisy areas~\cite{lee2019wide}. Consequently, error-only sampling frequently prioritizes high-error but low-leverage points, wasting gradient steps on updates with minimal global impact. It prompts us to dynamically select examples for fast convergence as described in the next subsection.

\subsection{NTK-Guided Implicit Neural Teaching}
\label{sec:nint}

\begin{algorithm}[t]
\caption{NTK-Guided Implicit Neural Teaching}
\label{alg:nint}
\textbf{Input:} Signal $S = \{(\mathbf{x}_i, \mathbf{y}_i)\}_{i=1}^N$, MLP $f_\theta$, batch size $B$, learning rate $\eta$, iterations $T$ 

\textbf{Output:} Trained INR $f_{\theta_T}$
\begin{algorithmic}[1]
\State Initialize $\theta_0$
\For{$t = 0$ to $T-1$}
    \State Forward pass: $\hat{\mathbf{y}}_i = f_{\theta_t}(\mathbf{x}_i)$ for all $i$
    \State Compute $\mathbf{g}^t = \big[ \nabla_f \mathcal{L}\bigl(f_{\theta^t}(\mathbf{x}_i), \mathbf{y}_i\bigr) \big]_{i=1}^N$
    \State Compute NTK row: $K_{\theta^t}(\mathbf{x}_i,:)$ for all $i$
    \State Select: $\mathcal{B}_t = \underset{\substack{\mathcal{B} \subseteq \{1,\dots ,N\} \\ |\mathcal{B}|=B}}{\arg\max}  \left\| \left[ K_{\theta^t}(\mathbf{x}_i,:) \cdot \mathbf{g}^t \right]_{i \in \mathcal{B}} \right\|_2$
    \State Update: $\theta_{t+1} \gets \theta_t -  \frac{\eta}{B} \sum_{i \in \mathcal{B}_t} \nabla_\theta \mathcal{L}(f_{\theta_t}(\mathbf{x}_i), \mathbf{y}_i)$
\EndFor
\State \Return $f_{\theta_T}$
\end{algorithmic}
\end{algorithm}

To overcome the isotropic-diagonal assumption of error-only sampling, we propose NTK-Guided Implicit Neural Teaching (NINT), a principled sampling strategy that explicitly accounts for the \textbf{heterogeneous self-leverage} and \textbf{functional coupling} encoded in the NTK. Rather than treating all high-error points equally, NINT prioritizes coordinates that are \textbf{both poorly fitted} and \textbf{globally influential}, \ie, those that drive large functional changes across the domain via strong self-leverage and cross-coordinate coupling.

Error-only sampling~\cite{gai2023egra,zhang2024nonparametric,zhang2025evos} selects a batch $\mathcal{B} \subseteq \{1,\dots,N\}$ of size $B$ by maximizing the norm of the loss gradient vector, \ie
\begin{equation*}
    \mathcal{B}^\star =\underset{|\mathcal{B}|=B}{\arg\max} \left\| \big[ \nabla_f \mathcal{L}\bigl(f_{\theta^t}(\mathbf{x}_i), \mathbf{y}_i\bigr) \big]_{i \in \mathcal{B}} \right\|_2.
\end{equation*}
This prioritizes points with large local prediction error (\ie, high output disparity between the current MLP prediction and the target signal). While some prior methods, such as EVOS~\cite{zhang2025evos}, augment this criterion by incorporating a Laplacian operator applied to both the predicted $f_{\theta^t}(\mathbf{x})$ and the given signal $\mathbf{y}$ to better capture high-frequency residuals, they still operate \emph{solely in the output space} and remain agnostic to the parameter-to-function mapping induced by the NTK. However, as highlighted in Sec.~\ref{sec:ntk}, this ignores the varying self-leverage and functional coupling encoded in the NTK, leading to inefficient updates that may prioritize high-error but low-impact points.

NINT, in contrast, selects samples to maximize the expected magnitude of the functional evolution, as captured by the NTK-augmented gradient. From Eq.~\ref{eq:func_evol}, the instantaneous change in the function $f_{\theta^t}(\mathbf{x})$ at any point $\mathbf{x}$ is governed by the product of the NTK row $K_{\theta^t}(\mathbf{x}_i, \mathbf{x})$ and the loss gradient at observed points. To accelerate global convergence, NINT prioritizes coordinates $\mathbf{x}_i$ that are \textit{both poorly fitted} and \textit{globally influential}, \ie, those inducing large functional shifts via \textbf{strong self-leverage} ($K_{\theta^t}(\mathbf{x}_i,\mathbf{x}_i)$) and \textbf{cross-coordinate coupling} ($K_{\theta^t}(\mathbf{x}_i,\mathbf{x}_j)$). Formally:
\begin{equation*}
    \mathcal{B}^\star =\underset{|\mathcal{B}|=B}{\arg\max} \left\| \left[ K_{\theta^t}(\mathbf{x}_i,:) \cdot \left[ \nabla_f \mathcal{L}\bigl(f_{\theta^t}(\mathbf{x}_j), \mathbf{y}_j\bigr) \right]_{j=1}^N \right]_{i \in \mathcal{B}} \right\|_2,
\end{equation*}
where $K_{\theta^t}(\mathbf{x}_i,:)\in \mathbb{R}^{1 \times N}$ denotes the row of the NTK matrix corresponding to $\mathbf{x}_i$. This incorporates both the local error magnitude and the coordinate's leverage on the broader function space. Pseudo code is in Algorithm \ref{alg:nint}.

\section{Results}
\label{sec:results}

\subsection{Experimental Setup}

\subsubsection{Settings and Metrics}
\label{sec:setup}

Following \cite{zhang2025evos,zhang2025expansive}, a portion $\xi$ of the training set is randomly selected. The remaining portion $1 - \xi$ is divided between NTK-guided sampling and error-based sampling. The NTK-guided ratio is $(1 - \xi) \exp(-\lambda t / \alpha)$, where $\lambda$ controls the decay rate, and $\alpha$ sets the NTK recomputation interval (reusing prior results otherwise). The error-based ratio constitutes the balance, selecting samples via fitting errors as in prior works~\cite{gai2023egra,zhang2024nonparametric}. 
We adopt this hybrid sampling to leverage NTK's global influence while transitioning to effective error-based selection, with hyperparameters tuned for computational balance (see ablations in Sec.~\ref{sec:hyper}).
Particularly, default hyperparameters are set as follows: $\xi=0.7$, $\alpha=10$, and $\lambda=1.0$.
For baselines, we used their official default settings. The default network is a 5$\times$256 SIREN \cite{strumpler2022implicit}, with further analysis of network size and architecture in Sec.~\ref{sec:size} and Sec.~\ref{sec:structure}.
All runs were on a single NVIDIA RTX 4090 GPU with 24 GB of memory. We measured reconstruction quality using PSNR, SSIM~\cite{wang2004image}, and LPIPS~\cite{zhang2018unreasonable}.

\subsubsection{Baseline Strategies and Datasets}

\begin{figure}[!t]
\centering
\includegraphics[width=0.45\textwidth]{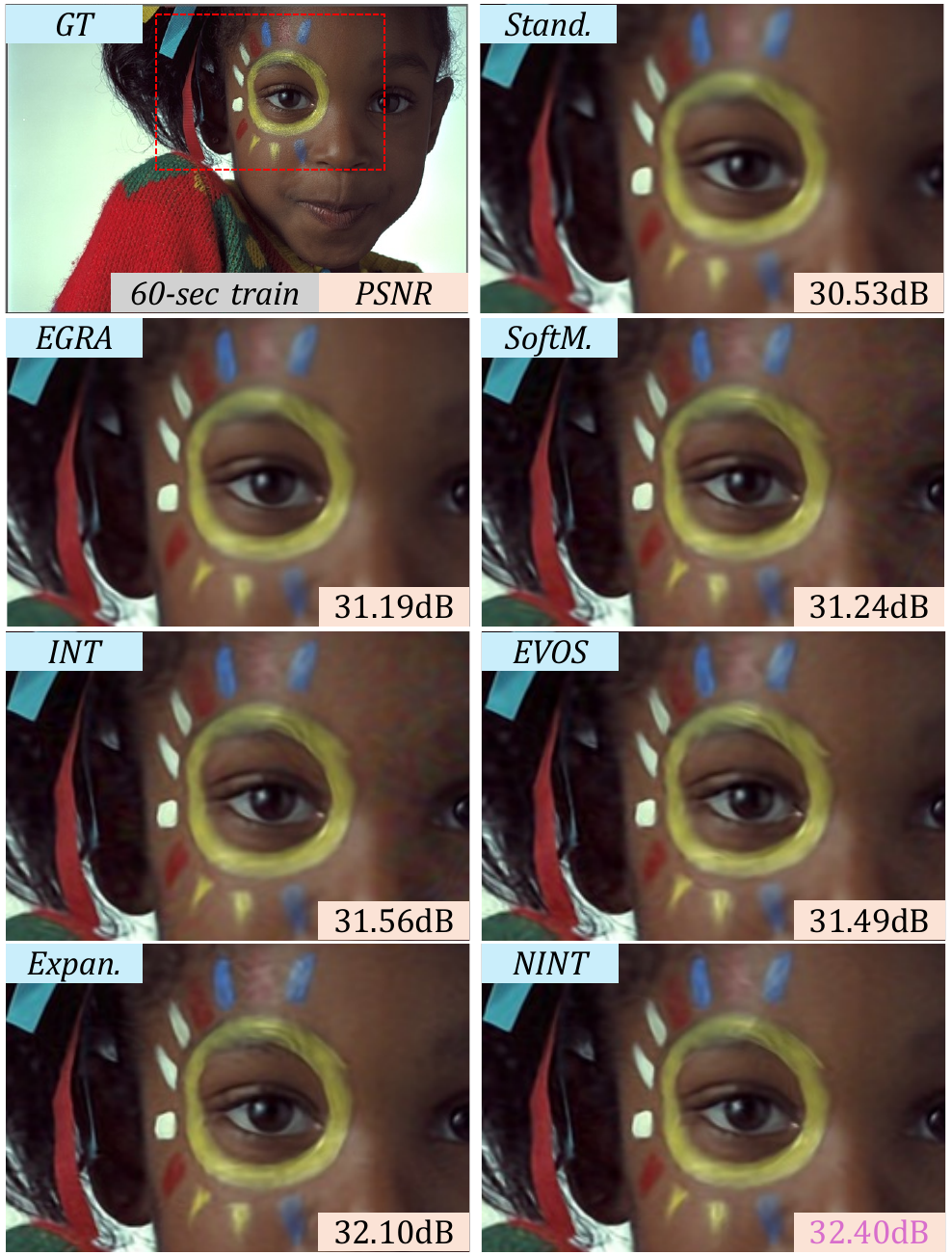}
\caption{Visual comparison of image reconstructions using different sampling strategies on the \textit{kodim15} image from the Kodak dataset~\cite{kodak}, after a fixed training duration of 60 seconds.}
\vspace{-3mm}
\label{fig:girl}
\end{figure}

We benchmark NINT against several leading INR acceleration methods: \textbf{(1)} Expansive Supervision (Expan.) \cite{zhang2025expansive}, \textbf{(2)} EVOS \cite{zhang2025evos}, \textbf{(3)} INT \cite{zhang2024nonparametric}, \textbf{(4)} SoftM. \cite{kheradmand2024accelerating}, \textbf{(5)} EGRA \cite{gai2023egra}, \textbf{(6)} Uniform Random Sampling (Unif.), and \textbf{(7)} conventional
full-coordinate training (Stand.). 
For EVOS \cite{zhang2025evos}, we incorporate the official variant, EVOS (w/o CFS.), where only vanilla L2 loss is implemented without Cross-Frequency Supervision; For Expan. \cite{zhang2025expansive}, to align with its official uniform random sampling ratio setting ($\xi=0.5$), we include a matching NINT variant for fair comparison.
All methods use a learning rate of $\eta=1\text{e}-4$ and a batch size of $B=20~\%$ of the full sample set (except $B=100~\%$ for Stand.).
Our 2D image experiments draw from the Kodak~\cite{kodak} and DIV2K~\cite{agustsson2017ntire} datasets. Furthermore, extended experiments on more images and other modalities, including 1D and 3D datasets, can be found in \textit{supplementary materials}.  

\subsection{Comparison with State-of-the-art Strategies}

Table~\ref{table:ite} presents the quantitative results under fixed training iteration budgets. NINT leads across all thresholds (250, 1000, and 5000 iterations) and metrics (PSNR, SSIM, LPIPS). This consistent superiority highlights the effectiveness of NTK-guided functional updates, which enable more informative sample selection and accelerate convergence. Notably, NINT not only achieves higher reconstruction quality at early stages, but also maintains its advantage as training progresses, indicating robust performance throughout the optimization process.

\begin{table*}[!t]
    \centering
    \begin{tabular}{l|ccc|ccc|ccc}
        \toprule
                & \multicolumn{3}{c|}{250 Iterations} & \multicolumn{3}{c|}{1000 Iterations} & \multicolumn{3}{c}{5000 Iterations} \\ 
        Strategy & PSNR$\uparrow$ &SSIM$\uparrow$ &LPIPS$\downarrow$ &  PSNR$\uparrow$ & SSIM$\uparrow$& LPIPS$\downarrow$ & PSNR$\uparrow$& SSIM$\uparrow$& LPIPS$\downarrow$\\
        \midrule
        \rowcolor{gray2}
        Stand. & 27.897 & 0.774	& 0.438	& 31.669&	0.845	&0.278	&39.763	&0.962	&0.022 \\
        \midrule
        Unif. & 27.660	&0.771	&0.443	&31.139	&0.836	&0.291	&37.137	&0.943	&0.069 \\
        EGRA \cite{gai2023egra} & 27.672	&0.772	&0.443	&31.243	&0.838	&0.289	&37.393	&0.945	&0.068\\
        SoftM. \cite{kheradmand2024accelerating}& 27.647	&0.747	&0.498	&31.381	&0.816	&0.338	&35.504&	0.920	&0.070 \\
        INT \cite{zhang2024nonparametric}& 27.568	&0.747	&0.499	&31.192	&0.815	&0.351	&\cellcolor{purple2}39.020	&0.943	&\cellcolor{purple2}0.035 \\
        EVOS \cite{zhang2025evos} (w/o CFS.) & 27.964	&0.753	&0.459	&31.657	&0.842	&0.260	&37.175	&0.935	&0.054 \\
        EVOS \cite{zhang2025evos} & 28.016	&0.755	&0.454	&31.719	&0.842	&0.262	&37.558	&0.940	&0.054 \\
        Expan. \cite{zhang2025expansive} ($\xi=0.5$)$^\dagger$	&28.034	&\cellcolor{purple2}0.773	&\cellcolor{purple2}0.420	&32.273	&\cellcolor{purple1}0.857	&\cellcolor{purple1}0.227	&38.350	&\cellcolor{purple2}0.948	&0.049\\
        Expan. \cite{zhang2025expansive} ($\xi=0.7$)	&27.990	&0.772	&0.426	&32.154	&\cellcolor{purple2}0.855	&0.240	&38.220	&0.947	&0.056\\
        NINT ($\xi=0.5$)	&\cellcolor{purple2}28.720	&0.770	&0.438	&\cellcolor{purple2}32.591	&0.851	&0.246	&38.082&	0.933	&0.037 \\
        NINT ($\xi=0.7$)	&\cellcolor{purple1}28.956&\cellcolor{purple1}0.776	&\cellcolor{purple1}0.414	&\cellcolor{purple1}32.640	&0.841	&\cellcolor{purple2}0.238	&\cellcolor{purple1}39.085	&\cellcolor{purple1}0.958	&\cellcolor{purple1}0.029     \\   
        
        \bottomrule
        \multicolumn{10}{l}{\scriptsize{$\dagger$ denotes the default setting in  Expan. \cite{zhang2025expansive}.}}
         
    \end{tabular}
    \vspace{-1mm}
    \caption
    {Performance metrics (PSNR↑, SSIM↑, LPIPS↓) at fixed training iterations (250, 1000, 5000) across various sampling strategies on image fitting tasks. \colorbox{purple1}{Purple}: the best performance; \colorbox{purple2}{Pale Purple}: the secondary performance.}
    \label{table:ite}
\end{table*}

\begin{table*}[!t]
    \centering
    \begin{tabular}{c|c|>{\columncolor{gray2}}c|cccccccccc}
\toprule
 & & \multicolumn{10}{c}{Strategy} \\
PSNR& Metric & Stand. & Uni. & EGRA & SoftM.& INT & EVOS & Expan.& Expan. & NINT  & NINT  \\
 &  & &  &  &  &  &   & ($\xi\!=\!0.5$)$^\dagger$ &  ($\xi\!=\!0.7$) &($\xi\!=\!0.5$) &  ($\xi\!=\!0.7$) \\
\midrule
\multirow{2}{*}{\centering 25} & iter$\downarrow$ & 80	&82	&82&	99	&100	&132	&90	&86	&\cellcolor{purple2}72	&\cellcolor{purple1}70  \\
& Time(s)$\downarrow$ & 8.40	&\cellcolor{purple2}5.94	&6.09	&7.03	&6.44	&8.90	&6.57	&6.34	&6.07	&\cellcolor{purple1}5.92 \\
\midrule
\multirow{2}{*}{\centering 30} & iter$\downarrow$ & 523	&626	&603	&589	&589	&490	&450	&454	&\cellcolor{purple1}380	&\cellcolor{purple2}384 \\
& Time(s)$\downarrow$& 49.11	&38.85	&37.85	&37.87	&33.01	&31.20	&28.59	&29.16	&\cellcolor{purple2}25.63	&\cellcolor{purple1}25.05 \\
\midrule
\multirow{2}{*}{\centering 35} & iter$\downarrow$ & 2043	&2760	&2619	&2624	&2023	&2302	&1802	&1988	&\cellcolor{purple2}1706	&\cellcolor{purple1}1644 \\
& Time(s)$\downarrow$ & 184.78	&168.46	&160.41	&165.42	&111.80	&143.20	&\cellcolor{purple2}111.01	&123.60	&108.90	&\cellcolor{purple1}102.88 \\
\bottomrule
\multicolumn{10}{l}{\scriptsize{$\dagger$ denotes the default setting in  Expan. \cite{zhang2025expansive}.}}
    \end{tabular}
    \vspace{-1mm}
    \caption
    {Iterations and runtime (in seconds) required to reach target PSNR thresholds across various sampling strategies on image fitting tasks. \colorbox{purple1}{Purple}: the best performance; \colorbox{purple2}{Pale Purple}: the secondary performance.}
    \label{table:psnr}
\vspace{-3mm}
\end{table*}

\begin{figure}[t]
\centering
\includegraphics[width=0.48\textwidth]{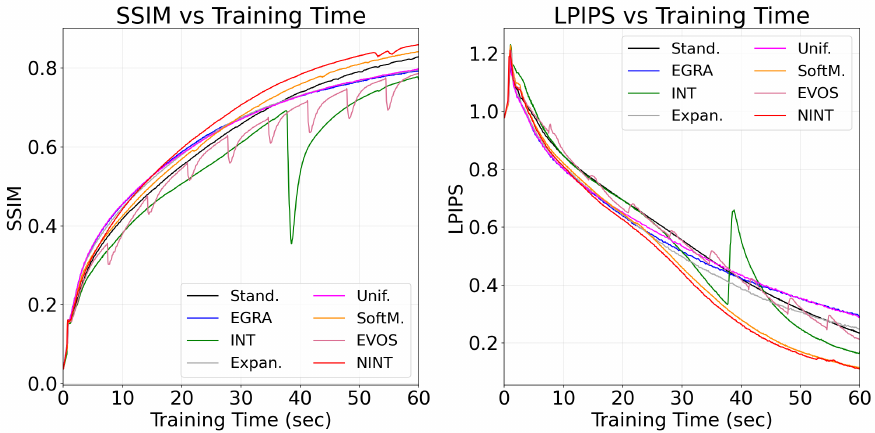}
\caption{Trends in SSIM~\cite{wang2004image} and LPIPS~\cite{zhang2018unreasonable} metrics over 60 seconds of training across various sampling strategies.}
\label{fig:dynam}
\vspace{-3mm}
\end{figure}

Table~\ref{table:psnr} examines efficiency by tracking iterations and time needed to hit specific PSNR targets. NINT demonstrates clear improvements in both training speed and resource utilization, outperforming all baselines across multiple target PSNR values ($25$, $30$, and $35$~dB). In particular, when compared to standard full-coordinate training (Stand.), NINT reduces the number of required iterations and total training time by up to $26.58~\%$ and \underline{$\mathbf{48.99}~\%$}, respectively. These savings make NINT ideal for INR tasks with tight time constraints, blending global and local sampling for faster, high-quality results.

Fig.~\ref{fig:girl} offers a visual comparison after 60 seconds of training.
Compared to baseline strategies, NINT produces crisper details, especially in the zoomed eye area, clearly preserving patterns and color boundaries of the facial painting. This aligns with its top PSNR of $32.40$~dB among all strategies, confirming that its quantitative advantage translates into meaningful visual improvements in INR tasks.

\begin{figure*}[!t]
\centering
\includegraphics[width=1.00\textwidth]{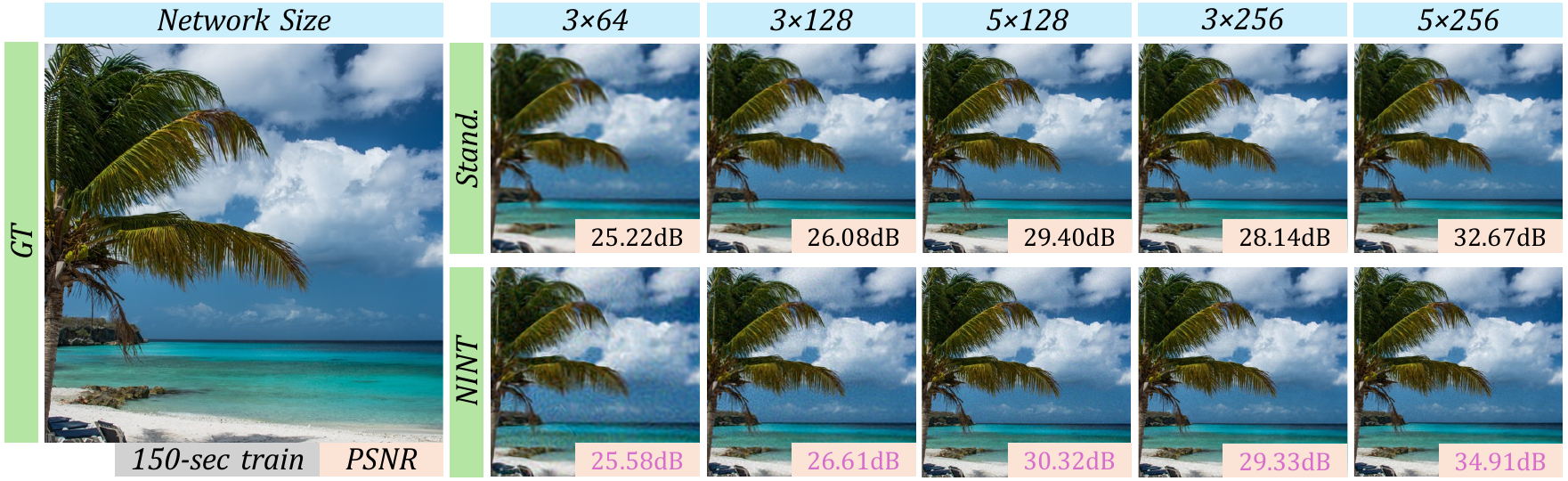}
\caption{Visual comparison of reconstruction quality across various network sizes, after training for 150 seconds on image \textit{07} from DIV2K~\cite{agustsson2017ntire}.}
\label{fig:size}
\vspace{-3mm}
\end{figure*}

Fig.~\ref{fig:dynam} tracks SSIM~\cite{wang2004image} and LPIPS~\cite{zhang2018unreasonable} over 60 seconds. 
Across both metrics and over the full training duration, NINT consistently achieves the highest structural and perceptual fidelity and outclasses all baselines. 
Notably, NINT demonstrates superior convergence behavior, attaining higher SSIM and lower LPIPS values faster in training and maintaining this advantage through the final iteration.
This indicates that NINT not only yields better image reconstruction quality but also does so more efficiently, underscoring its superior effectiveness.

\subsection{Ablation Studies}

\subsubsection{Adaptability for Network Sizes}
\label{sec:size}

\begin{table}[t]
    \centering
    \begin{tabular}{l|ccc|c}
\toprule
Network Size &  \multicolumn{3}{c|}{PSNR $\uparrow$} & Time (s) $\downarrow$ \\
& 500 & 1000 & 2500 &  \\
\midrule
1$\times$64 & 11.835 & 17.137 & 18.305 & N/A \\
1$\times$64 (NINT) & \cellcolor{purple1}13.890 & \cellcolor{purple1}17.606 & \cellcolor{purple1}18.527 & N/A \\
\midrule
3$\times$64 & \cellcolor{purple1}22.291 & 23.527 & 25.110 & 159.04 \\
3$\times$64 (NINT) & 22.023 & \cellcolor{purple1}23.611 & \cellcolor{purple1}25.257 & \cellcolor{purple1}141.14 \\
\midrule
3$\times$128 & 23.172 & 24.169 & 26.140 & 92.16 \\
3$\times$128 (NINT) & \cellcolor{purple1}23.195 & \cellcolor{purple1}24.506 & \cellcolor{purple1}26.521 & \cellcolor{purple1}72.14 \\
\midrule
5$\times$128 & 24.486 & 26.036 & 30.114 & 45.34 \\
5$\times$128 (NINT) & \cellcolor{purple1}24.948 & \cellcolor{purple1}27.363 & \cellcolor{purple1}30.259 & \cellcolor{purple1}31.47 \\
\midrule
3$\times$256 & 24.075 & 25.785 & 28.852 & 57.68 \\
3$\times$256 (NINT) & \cellcolor{purple1}24.404 & \cellcolor{purple1}27.761 & \cellcolor{purple1}29.444 & \cellcolor{purple1}39.75 \\
\midrule
5$\times$256 & 25.613 & 28.685 & 33.693 & 35.42 \\
5$\times$256 (NINT) & \cellcolor{purple1}26.851 & \cellcolor{purple1}31.274 & \cellcolor{purple1}35.100 & \cellcolor{purple1}22.16 \\
\bottomrule
    \end{tabular}
    \caption
    {Comparison of PSNR at fixed training iterations (500, 1000, 2500) and elapsed time for 3000-iteration training across different network sizes, with and without NINT. Better performance is marked as \colorbox{purple1}{purple}.
    }
    \label{table:layer}
    \vspace{-3mm}
\end{table}

Since INR tasks differ in complexity, sampling strategies must adapt well. We tested NINT across SIREN~\cite{strumpler2022implicit} networks from 1$\times$64 to 5$\times$256 to check its fit for varying capacities.
Specifically, we compare: \textbf{1)} PSNR at fixed training iterations including 500, 1000 and 2500; \textbf{2)} the elapsed time for reaching the PSNR~$=25$~dB threshold.
As Table \ref{table:layer} indicates, NINT's time savings grow with network size, hitting \underline{$37.44~\%$} at 5$\times$256. Even for the small 1$\times$64 network, which cannot always hit the threshold (marked as N/A) due to its limited capacity, NINT still brings up to $17.36~\%$ PSNR improvement at fixed iteration checkpoints under such setting.  
Furthermore, NINT demonstrates consistent superiority of PSNR at given training iterations, which clearly confirms the effectiveness of incorporating NTK-guided sampling for training acceleration while satisfying distinct expressiveness requirements of INR tasks.

\begin{table}[t]
    \centering
    \begin{tabular}{l|ccc|c}
\toprule
Network Structure & \multicolumn{3}{c|}{PSNR $\uparrow$} & Time (s) $\downarrow$ \\
& 20s & 60s & 120s &  \\
\midrule
MLP & 14.91 & 18.13 & 20.67 & 374.64 \\
MLP + NINT & \cellcolor{purple1}15.11 & \cellcolor{purple1}19.72 & \cellcolor{purple1}21.68 & \cellcolor{purple1}266.32 \\
\midrule
FFN \cite{tancik2020fourier} & 16.75 & 26.9 & 31.44 & 54.19 \\
FFN + NINT & \cellcolor{purple1}18.88 & \cellcolor{purple1}27.39 & \cellcolor{purple1}31.48 & \cellcolor{purple1}48.75 \\
\midrule
FINER \cite{liu2023finer} & 28.12 & 31.15 & 33.01 & 6.90 \\
FINER + NINT & \cellcolor{purple1}29.54 & \cellcolor{purple1}31.28 & \cellcolor{purple1}35.61 & \cellcolor{purple1}6.76 \\
\midrule
GAUSS \cite{ramasinghe2022beyond} & \cellcolor{purple1}21.51 & 25.87 & 33.05 & 56.23 \\
GAUSS + NINT & 21.46 & \cellcolor{purple1}27.11 & \cellcolor{purple1}33.38 & \cellcolor{purple1}48.24 \\
\midrule
PEMLP \cite{tancik2020fourier} & 18.41 & 25.09 & 29.11 & 59.02 \\
PEMLP + NINT & \cellcolor{purple1}19.95 & \cellcolor{purple1}25.91 & \cellcolor{purple1}30.45 & \cellcolor{purple1}49.13 \\
\midrule
SIREN \cite{strumpler2022implicit} & 27.45 & 30.51 & 32.44 & 8.25 \\
SIREN + NINT & \cellcolor{purple1}29.28 & \cellcolor{purple1}32.40 & \cellcolor{purple1}35.47 & \cellcolor{purple1}5.81 \\
\midrule
WIRE \cite{saragadam2023wire} & 16.88 & 23.86 & 27.17 & 83.30 \\
WIRE + NINT & \cellcolor{purple1}17.62 & \cellcolor{purple1}26.62 & \cellcolor{purple1}29.13 & \cellcolor{purple1}47.23 \\
\bottomrule
    \end{tabular}
    \caption
    {Comparison of PSNR at fixed training times (20s, 60s, 120s) and elapsed time to reach PSNR=25 threshold across various INR network structures, with and without NINT. Better performance is highlighted in \colorbox{purple1}{purple}.
    }
    \label{table:structure}
    \vspace{-3mm}
\end{table}

\begin{table*}[t!]
    \centering
    \begin{tabular}{l|cc|cc|cc|cc|cc|cc}
\toprule
\multirow{3}{*}{Setting} & \multicolumn{6}{c|}{Iterations $\downarrow$} & \multicolumn{6}{c}{Time (s) $\downarrow$}  \\
& \multicolumn{2}{c|}{PSNR} &\multicolumn{2}{c|}{SSIM}&\multicolumn{2}{c|}{LPIPS} & \multicolumn{2}{c|}{PSNR} &\multicolumn{2}{c|}{SSIM}&\multicolumn{2}{c}{LPIPS}  \\
& 30 & 35 & 0.8 & 0.9 &0.2&0.1& 30 & 35 & 0.8 & 0.9&0.2&0.1  \\

\midrule
NINT (default) &389&	\cellcolor{purple1}1644&	399&\cellcolor{purple1}	1639&1172	&2082&\cellcolor{purple1}25.09 &	102.88&\cellcolor{purple1}	26.00&	102.56&73.51&	130.21\\

\midrule
$\alpha\!=\!$~1  &\cellcolor{pink}411&	1682&\cellcolor{pink}	424&	1685 &1258	&2297& 27.08&	107.43&	27.82&	107.65&80.42&146.21\\
$\alpha\!=\!$~20   &401&    1698&	400&	1717 &\cellcolor{pink}1338&	1941\cellcolor{purple1}&26.19&	106.24&	26.15&	107.44&83.74&\cellcolor{purple1}121.31\\
$\alpha\!=\!$~50 &383&	1650&\cellcolor{purple1}	394&	1651 &\cellcolor{purple1}1155&	2384&25.39&	104.07&	26.08&	104.12&\cellcolor{purple1}73.09&	149.60\\

\midrule
$\xi\!=\!$~0.4&391&	1659&	431&	1832&1327&	2001&26.03	&104.55	&28.46	&115.36&\cellcolor{pink}83.77&	129.41\\
$\xi\!=\!$~0.5 &380&	1706&	414&	1762&1200&	1973&25.63	&108.89	&27.65	&112.44&76.98&	125.69\\
$\xi\!=\!$~0.6&\cellcolor{purple1}376&	1656&	401&	1693&1287&\cellcolor{pink}	2491&24.71	&103.50	&26.22	&105.80&80.51&\cellcolor{pink}	155.08\\
$\xi\!=\!$~0.8&400&	1647&	399&	1644&1181&	2331&26.16&	102.95&	26.04&\cellcolor{pink}	120.73&74.04&	145.26\\

\midrule
$\lambda\!=\!$~0.1&390&	1700&	404&	1711&1202&	1962&\cellcolor{pink}28.48&\cellcolor{pink}	110.37&\cellcolor{pink}	29.37&	110.99&79.42&	126.50\\
$\lambda\!=\!$~0.5&394&\cellcolor{pink}	1731&	402&\cellcolor{pink}	1742&1239&	1951&26.28	&109.43	&26.79	&110.10&78.68&	123.02\\
$\lambda\!=\!$~2.0&380&	1648&	404&	1643&1208&	2028&24.82	&\cellcolor{purple1}102.85	&26.22	&\cellcolor{purple1}102.55&75.57&	126.24\\  
\bottomrule
    \end{tabular}
    \caption
    {Ablation study on the impact of hyperparameters $\alpha$, $\xi$, and $\lambda$ in NINT, showing iterations and runtime (in seconds) required to reach target thresholds for PSNR, SSIM, and LPIPS on image fitting tasks.
    \colorbox{purple1}{Purple}: the best performance;
    \colorbox{pink}{Orange}: the worst performance.}
    \label{table:para}
    \vspace{-3mm}
\end{table*}

Fig. \ref{fig:size} presents a visual comparison of network sizes under a fixed 150‑second training budget.
While both strategies benefit from increased capacity, NINT consistently outperforms Stand., with PSNR improving from $25.58$~dB (3$\times$64) to $34.91$~dB (5$\times$256). Notably, the performance gap widens as the network scales, indicating that NINT more effectively leverages larger capacity for higher-fidelity reconstructions while maintaining its advantage even with limited model capacity. This highlights NINT’s scalability and robustness across model sizes.

\subsubsection{Adaptability for Network Structures}
\label{sec:structure}

To probe NINT's adaptability for network structures, we conduct experiments across a diverse set of neural architectures, including plain MLP and its extensive variants: FFN \cite{tancik2020fourier}, FINER \cite{liu2023finer}, GAUSS \cite{ramasinghe2022beyond}, PEMLP \cite{tancik2020fourier}, SIREN \cite{strumpler2022implicit}, and WIRE \cite{saragadam2023wire}. These architectures, characterized by heterogeneous designs in frequency-induced encoding and activation functions, encompass a broad spectrum of INR scenarios and pose unique challenges for sampling strategy robustness. 
As summarized in Table~\ref{table:structure}, we evaluate performance by comparing PSNR at fixed training times (20, 60, and 120 seconds), as well as the time required to reach a PSNR of $25$~dB. The results demonstrate that NINT consistently delivers superior performance across all tested structures, reducing training time by up to \underline{$43.30~\%$} and improving PSNR by as much as $11.57~\%$ under comparable training budgets. These improvements are particularly notable in architectures with complex frequency encoding, where traditional sampling strategies often struggle to efficiently allocate computational resources.
By consistently delivering acceleration and quality improvements regardless of model type, NINT is well-suited for practical INR applications with diverse network choices.

\subsubsection{Adaptability for NINT Settings}
\label{sec:hyper}

We further assess the sensitivity of NINT to its hyperparameters across multiple metrics, considering both iteration- and time-based efficiency. As shown in Table~\ref{table:para}, NINT achieves robust performance under the default configuration (see Sec.~\ref{sec:setup}), consistently ranking at or near the top across all evaluation criteria (\eg, fastest time to PSNR~$=30$~dB: 25.09 seconds; lowest time to SSIM~$=0.8$: 26.00 seconds). Importantly, NINT maintains strong performance even when deviating from default settings, demonstrating flexibility in optimization: for instance, $\alpha\!=\!50$ yields the best time for LPIPS~$=0.2$, $\xi\!=\!0.6$ achieves the lowest iteration count for PSNR~$=30$~dB, and $\lambda\!=\!2.0$ attains the fastest time for PSNR~$=35$~dB and SSIM~$=0.9$. Even in worst-case scenarios (e.g., $\alpha\!=\!1$ or $\lambda\!=\!0.1$), NINT’s results remain close to optimal, underscoring its stability across diverse configurations. Overall, these findings highlight that NINT delivers plug-and-play efficiency and robust performance without the need for intricate hyperparameter tuning.

\section{Concluding Remarks and Future Work}
\label{sec:conclusion}

In this work, we introduced NINT, a sampling-based strategy that accelerate INR training by leveraging NTK to dynamically select coordinates maximizing global functional updates, integrating fitting errors with insights into heterogeneous self-leverage and cross-coordinate coupling to overcome the inefficiencies of error-only methods that ignore the NTK's off-diagonal elements and lead to suboptimal progress. Our NTK-centric analysis revealed these limitations, while NINT prioritizes NTK-augmented gradient norms for faster convergence. Extensive benchmarks show NINT outperforming baselines in PSNR, SSIM, and LPIPS, reducing time and iterations to quality thresholds by up to $49~\%$ and $27~\%$ compared to full-batch training, with visuals confirming sharper details achieved faster. It scales effectively with network size, yielding greater savings on larger models, adapts seamlessly across diverse architectures with up to $43.3~\%$ runtime cuts, and demonstrates robustness to hyperparameter variations, ensuring reliable performance without extensive tuning. Overall, NINT enhances INR practicality by speeding up training without architectural modifications or additional data, surpassing recent samplers~\cite{gai2023egra,zhang2024nonparametric,zhang2025expansive,zhang2025evos,kheradmand2024accelerating}. Future work includes NTK approximations to lower overhead, adaptive batching, and integration with hybrid architectures for applications in neural signal processing.
\section*{Acknowledgments}
This work was supported in part by the Theme-based Research Scheme (TRS) project T45-701/22-R of the Research Grants Council of Hong Kong, and in part by the AVNET-HKU Emerging Microelectronics and Ubiquitous Systems (EMUS) Lab.
{
    \small
    \bibliographystyle{ieeenat_fullname}
    \bibliography{main}
}

\clearpage
\setcounter{page}{1}
\maketitlesupplementary

%

\section{1D Audio Fitting Task}

\subsection{Background and Settings} 
\label{sec:1d}
We represent raw 1D audio waveforms as a continuous function $F_\theta: \mathbb{R} \to \mathbb{R}$ that maps time $t$ to its instantaneous amplitude $a = F_\theta(t)$. 
For this task, we use the \textit{test.clean} split of the LibriSpeech~\cite{panayotov2015librispeech} dataset.
The implicit representation is a 5$\times$256 SIREN \cite{strumpler2022implicit}. For NINT, we set the hyperparameters $\xi=0.7$, $\alpha=10$, and $\lambda=1.0$. All strategies share the learning rate of $\eta=1\text{e}-4$ and the batch size of $B=20~\%$ (except $B=100~\%$ for Stand.). Since EGRA~\cite{gai2023egra}, SoftM.~\cite{kheradmand2024accelerating}, and Expan.~\cite{zhang2025expansive} are born with incompatible designs for 1D audio fitting task, we compare NINT with Stand., Unif., INT~\cite{zhang2024nonparametric}, and EVOS~\cite{zhang2025evos} (with its crossover component disabled for fair comparison).
Moreover, we introduce the step-wise batch size scheduler from INT~\cite{zhang2024nonparametric} to produce variant strategies, which increases $B$ as training iteration grows.
For metrics, we adopt SI-SNR~\cite{le2019sdr}, STOI~\cite{taal2011algorithm}, PESQ~\cite{rix2001perceptual} to evaluate audio reconstruction qualities.

\subsection{Experimental Results} 

\textbf{Quantitative Analysis.}
As shown in Table~\ref{table:audio}, NINT consistently achieves the best or near-best performance across all training durations and evaluation metrics. Specifically, after 3 seconds of training, where baseline strategies still produce highly degraded outputs (\eg, SI-SNR: INT\,=\,$0.97$~dB; EVOS\,=\,$2.18$~dB), NINT clearly surpasses others by large margins. 
Applying the batch-size scheduler, all strategies generally improves SI-SNR and PESQ, but can slightly reduce STOI in some cases (\eg, \underline{Unif.}).
\underline{NINT}, however, benefits consistently from the scheduler without such degradation, implying its sampling distribution is more resilient to optimization dynamics shifts.
These results confirm NINT’s ability for rapid convergence in audio fitting tasks.

\begin{figure}[!t]
\centering
\includegraphics[width=0.49\textwidth]{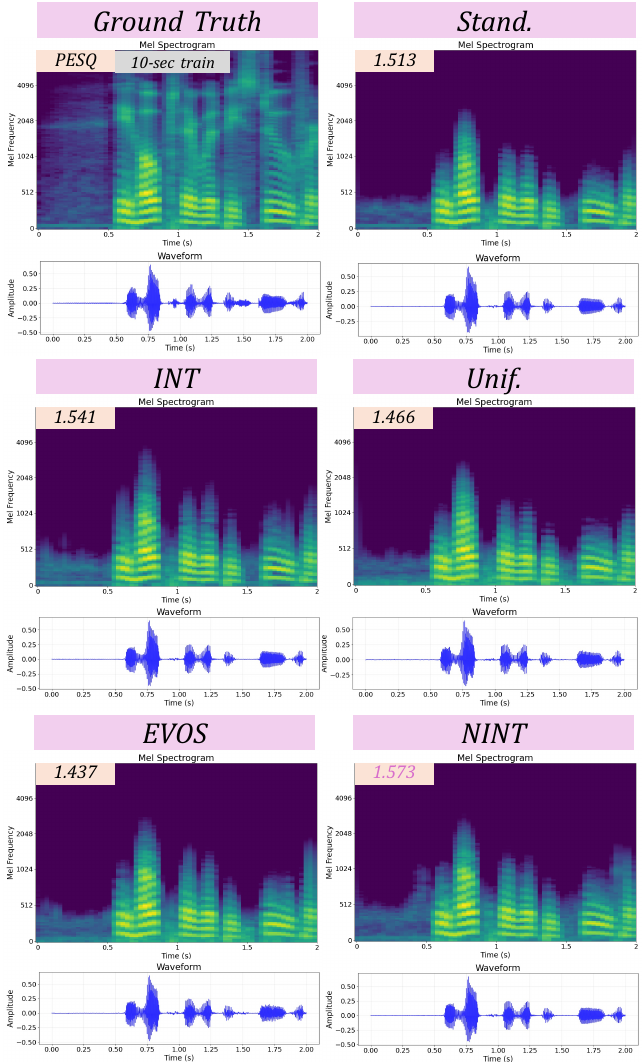}
\caption{Visual comparison (Mel spectrogram and waveform) of audio reconstructions using different sampling strategies on the \textit{test.clean} split of Librispeech~\cite{panayotov2015librispeech} dataset, after a fixed training duration of 10 seconds.}
\label{fig:audio}
\end{figure}

\noindent\textbf{Qualitative Analysis.}
Fig.~\ref{fig:audio} provides qualitative validation for the 10-second regime. The spectrogram and waveform of Unif. and EVOS show blurred harmonic structures and phase inconsistencies, while NINT preserves fine-grained harmonic structures and high-frequency energy (\eg, over 3 kHz), corroborating its numerical results of top PESQ and SI-SNR scores.
In summary, NINT establishes a new state-of-the-art strategy for 1D audio fitting under time-constrained training, delivering both fast convergence and high perceptual quality, validated quantitatively and qualitatively.

\begin{table*}[t!] 
    \centering
    \begin{tabular}{l|ccc|ccc|ccc}
        \toprule
                & \multicolumn{3}{c|}{3 seconds} & \multicolumn{3}{c|}{10 seconds} & \multicolumn{3}{c}{15 seconds} \\ 
        Strategy & SI-SNR$\uparrow$ &STOI$\uparrow$ &PESQ$\uparrow$ & SI-SNR$\uparrow$ &STOI$\uparrow$ &PESQ$\uparrow$ & SI-SNR$\uparrow$ &STOI$\uparrow$ &PESQ$\uparrow$\\
        \midrule
        \rowcolor{gray2}
        Stand. & -4.786	&	0.507	&	1.098	&	13.054	&	0.698	&	1.513	&	14.455	&	0.703	&	1.570 \\
        \midrule
        Unif. & -4.281	&	0.501	&	1.099	&	11.440	&\cellcolor{purple1}0.704	&	1.466	&	12.472	&	0.705	&	1.493 \\
        INT \cite{zhang2024nonparametric}&0.968	&	0.590	&	1.175	&	14.053	&	0.694	&	1.541	&\cellcolor{purple1}14.948	&	0.714	&	1.675 \\
        EVOS \cite{zhang2025evos} & 2.178	&	0.563	&	1.107	&	12.014	&	0.697	&	1.437	&	13.438	&	0.695	&	1.547 \\
        NINT &\cellcolor{purple1}3.842	&\cellcolor{purple1}0.614	&\cellcolor{purple1}1.200	&\cellcolor{purple1}14.280	&	0.702	&\cellcolor{purple1}1.573	&	14.803	&\cellcolor{purple1}0.715	&\cellcolor{purple1}1.834 \\
    
        \midrule
        \underline{Unif.} &-3.668	&	0.544	&	1.176	&	11.246	&\cellcolor{purple1}0.701	&	1.430	&	13.527	&	0.708	&	1.517 \\
        \underline{INT} \cite{zhang2024nonparametric}&1.919	&	0.605	&	1.214	&	14.201	&	0.695	&	1.603	&	14.584	&	0.715	&	1.751 \\
        \underline{EVOS} \cite{zhang2025evos} & 2.371	&	0.618	&	1.138	&	12.680	&	0.695	&	1.490	&	12.999	&	0.716	&	1.681 \\ 
        \underline{NINT} &\cellcolor{purple1}4.988	&\cellcolor{purple1}0.637	&\cellcolor{purple1}1.217	&\cellcolor{purple1}14.286	&\cellcolor{purple1}0.701	&\cellcolor{purple1}1.609	&\cellcolor{purple1}14.977	&\cellcolor{purple1}0.719	&\cellcolor{purple1}1.836 \\
        
        \bottomrule
        \multicolumn{10}{l}{\scriptsize{\underline{Underline} denotes step-wise batch size scheduler in INT~\cite{zhang2024nonparametric}.}}
    \end{tabular}
    \caption{Performance metrics (SI-SNR$\uparrow$, STOI$\uparrow$, PESQ$\uparrow$) at fixed training times across various sampling strategies on 1D audio fitting tasks. \colorbox{purple1}{Purple} denotes the best performance.}
    \label{table:audio}
\end{table*}

\begin{table*}[t] 
    \centering
    \begin{tabular}{l|cc|cc|cc|cc|cc}
        \toprule
        \multirow{3}{*}{Strategy} & \multicolumn{2}{c|}{500 Iters} & \multicolumn{2}{c|}{1k Iters} & \multicolumn{2}{c|}{2k Iters} &\multicolumn{2}{c|}{5k Iters} &\multicolumn{2}{c}{10k Iters}\\ 
        & IoU$\uparrow$&	CHD$\downarrow$	&IoU$\uparrow$	&	CHD$\downarrow$	&IoU$\uparrow$	&	CHD$\downarrow$	&IoU$\uparrow$&	CHD$\downarrow$	&IoU$\uparrow$&	CHD$\downarrow$\\
        &&($\times$1e-3)&&($\times$1e-3)&&($\times$1e-3)&&($\times$1e-3)&&($\times$1e-3)\\
        \midrule
        \rowcolor{gray2}
        Stand. & 0.9545	&	6.353	&	0.9610	&	6.206	&	0.9681	&	6.106	&	0.9776	&	5.975	&	0.9811	&	5.942 \\
        \midrule
        Unif. &0.9434	&	6.636	&	0.9584	&	6.235	&	0.9629	&	6.139	&	0.9733	&	6.023	&	0.9801	&	\cellcolor{purple1}5.978 \\
        INT \cite{zhang2024nonparametric}&0.9483	&	6.520	&	0.9594	&	6.623	&	0.9665	&	6.153	&	0.9749	&	6.111	&	0.9805	&	6.062 \\
        EVOS \cite{zhang2025evos} & 0.9538	&	\cellcolor{purple1}6.405	&	0.9593	&	\cellcolor{purple1}6.218	&	0.9640	&	6.123	&	0.9728	&	6.070	&	0.9802	&	6.019 \\
        NINT &\cellcolor{purple1}0.9562	&	6.408	&	\cellcolor{purple1}0.9630	&	6.221	&	\cellcolor{purple1}0.9666	&	\cellcolor{purple1}6.116	&	\cellcolor{purple1}0.9762	&	\cellcolor{purple1}6.023	&	\cellcolor{purple1}0.9817	&	\cellcolor{purple1}5.978 \\
    
        \midrule
        \underline{Unif.} &0.9436	&	6.636	&	0.9587	&	6.232	&	0.9640	&	\cellcolor{purple1}6.135	&	0.9740	&	6.016	&	0.9811	&	5.958 \\
        \underline{INT} \cite{zhang2024nonparametric}&0.9483	&	6.518	&	0.9595	&	6.622	&	0.9631	&	6.150	&	0.9751	&	6.100	&	0.9805	&	6.020 \\
        \underline{EVOS} \cite{zhang2025evos} &0.9540	&	6.404	&	0.9596	&	6.219	&	0.9644	&	6.149	&	0.9733	&	6.059	&	0.9814	&	5.999\\ 
        \underline{NINT} &\cellcolor{purple1}0.9563	&	\cellcolor{purple1}6.391	&	\cellcolor{purple1}0.9637	&	\cellcolor{purple1}6.216	&	\cellcolor{purple1}0.9670	&	6.138	&	\cellcolor{purple1}0.9770	&	\cellcolor{purple1}6.005	&	\cellcolor{purple1}0.9825	&	\cellcolor{purple1}5.921\\
        
        \bottomrule
        \multicolumn{10}{l}{\scriptsize{\underline{Underline} denotes step-wise batch size scheduler in INT~\cite{zhang2024nonparametric}.}}
    \end{tabular}
    \caption{Performance metrics (IoU$\uparrow$ and CHD$\downarrow$) at fixed training iterations across various sampling strategies on 3D shape fitting tasks. \colorbox{purple1}{Purple} denotes the best performance.}
    \label{table:3d}
\end{table*}

\section{3D Shape Fitting Task}
\subsection{Backgound and Settings}

To encode 3D shapes, we employed Signed Distance Fields (SDF), an established method in computer graphics~\cite{jones20063d}.
The objective is to learn a mapping $F_\theta(\mathbf{x})$ that takes a point’s coordinates $(x, y, z)$ and outputs $s$, the signed distance between that point and the object’s surface.
Network configuration, NINT hyperparamters, and learning rate follow Sec. \ref{sec:1d}, while all strategies share the batch size of $B=40~\%$ (except $B=100~\%$ for Stand.).
Similar to incompatibility issues as illustrated in Sec. \ref{sec:1d}, we compare NINT with Stand., Unif., INT~\cite{zhang2024nonparametric}, and EVOS~\cite{zhang2025evos} (crossover component disabled for compatibility), with step-wise batch size scheduler from INT~\cite{zhang2024nonparametric} introduced to produce variants. 
Evaluation are performed on the Stanford 3D Scanning Repository~\cite{stanford3d}, a well-known dataset which provides high-quality 3D scans of real-world objects and is widely used for research in computer graphics.
For training set, 50,000 points are randomly sampled from the 3D shape surface on both coarse (Laplacian noise with variance 0.1) level and fine (Laplacian noise with variance 0.001) level to constitute the training set for each iteration following~\cite{lindell2022bacon}.
For metrics, we evaluate 3D reconstruction quality using: Intersection over Union (IoU), the volumetric overlap between predicted and ground-truth occupancies as a measure of global shape accuracy; and the Chamfer Distance (CHD), which is the bidirectional mean squared distance between surface point samples.

\begin{figure*}[!t]
\centering
\includegraphics[width=0.90\textwidth]{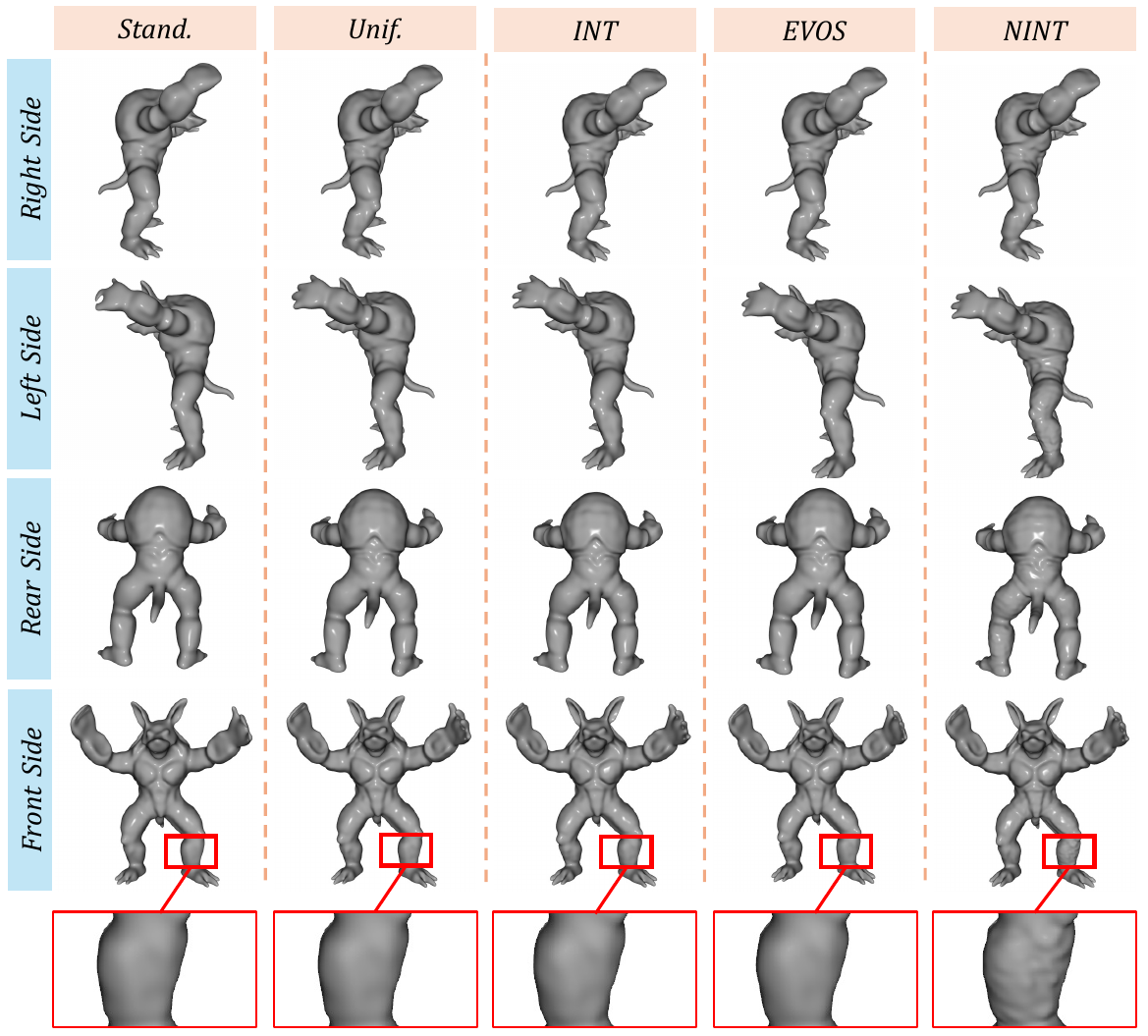}
\caption{Visual comparison of 3D shape reconstructions using different sampling strategies on the \textit{Armadillo} of Stanford 3D Scanning Repository~\cite{stanford3d} dataset, after a fixed training duration of 10 minutes.}
\label{fig:3d}
\end{figure*}

\subsection{Experimental Results}

\textbf{Quantitative Analysis.}
As shown in Table~\ref{table:3d}, NINT consistently achieves the best or near-best performance across training budgets and metrics.
Remarkably, NINT attains the highest IoU at every iteration threshold (500, 1k, 2k, 5k, and 10k) and the best CHD in most cases, underscoring its robust sample efficiency and balanced optimization of both global occupancy (IoU) and surface precision (CHD).
Notably, when combined with the step-wise batch size scheduler, NINT shows consistent improvement over its vanilla counterpart (\eg, boosting IoU from $0.9817$ to $0.9825$) and reducing CHD from $5.978~\text{e}-3$ to $5.921~\text{e}-3$ at 10k iterations which represents the best overall results.
In contrast, while schedulers benefit some baselines (\eg, \underline{Unif.} achieves CHD \,=\, $5.958~\text{e}-3$), they often fail to improve both metrics simultaneously (\eg, \underline{INT} improves CHD slightly but lags in IoU).
These results confirm that NTK-guided sampling can effectively prioritize informative regions (\eg, near-surface points with high gradient variance), accelerating convergence without sacrificing geometric fidelity, making it well-suited for 3D SDF learning tasks.

\noindent\textbf{Qualitative Analysis.}
Fig.~\ref{fig:3d} shows a visualized result of 10-minute 3D shape training on the \textit{Armadillo} of Stanford 3D Scanning Repository~\cite{stanford3d}.
Particularly, Stand. and Unif. exhibit noticeable surface noise and missing thin structures; INT and EVOS show improved topology but still suffer from bulging or over-smoothed regions. NINT, however, produces clean, high-fidelity reconstructions with sharp edges, accurate thin parts, and minimal artifacts.
This corroborates the superiority of NINT's top numerical results (outstanding IoU and CHD scores), confirming that NTK's guidance effectively preserves both global shape coherence and local surface details.

\section{Extended Evaluation on 2D Images}

\begin{figure*}[!t]
\centering
\includegraphics[width=0.9\textwidth]{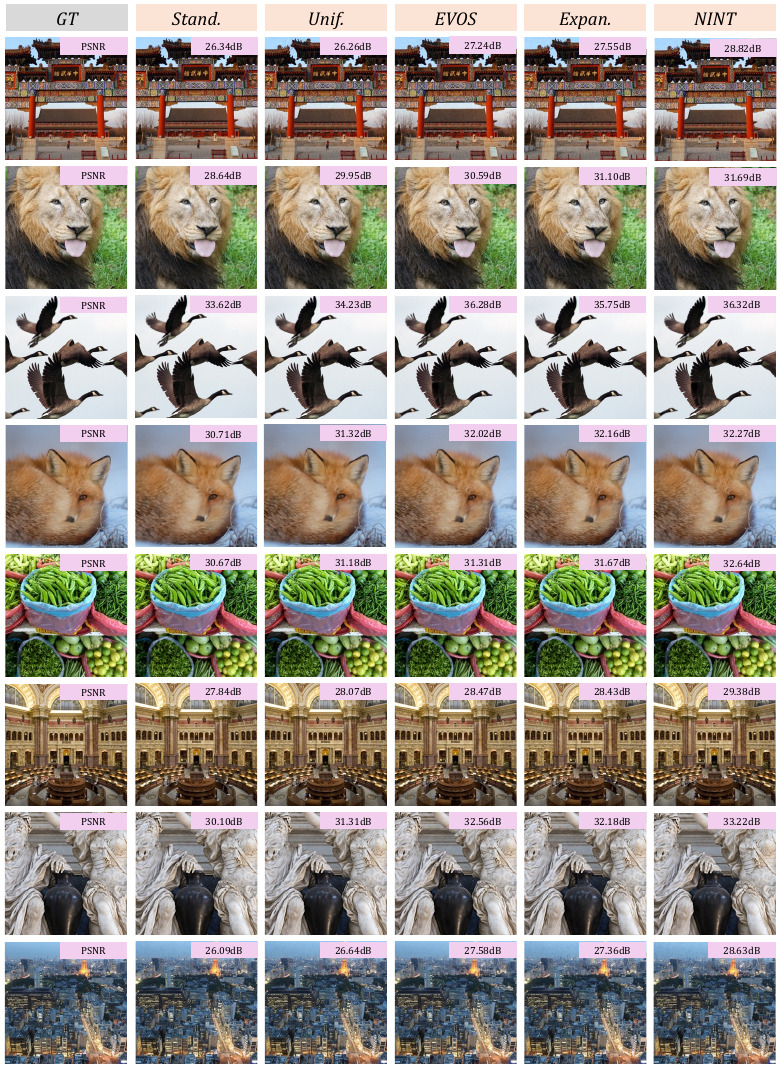}
\caption{Visual comparison of 2D image reconstructions using different sampling strategies on DIV2K~\cite{agustsson2017ntire} dataset, after a fixed training duration of 60 seconds.}
\vspace{-2mm}
\label{fig:div2k}
\end{figure*}

\begin{table*}[t!]
    \centering
    \begin{tabular}{l|cc|cc|cc|cc|cc|cc}
\toprule
\multirow{3}{*}{Setting} & \multicolumn{6}{c|}{Iterations $\downarrow$} & \multicolumn{6}{c}{Time (s) $\downarrow$}  \\
& \multicolumn{2}{c|}{PSNR} &\multicolumn{2}{c|}{SSIM}&\multicolumn{2}{c|}{LPIPS} & \multicolumn{2}{c|}{PSNR} &\multicolumn{2}{c|}{SSIM}&\multicolumn{2}{c}{LPIPS}  \\
& 30 & 35 & 0.8 & 0.9 &0.2&0.1& 30 & 35 & 0.8 & 0.9&0.2&0.1  \\

\midrule
\rowcolor{gray2}
Stand. &523	&	2043	&	420	&	1871	&	1645	&	2454	&	49.11	&	184.78	&	39.00	&	169.23	&	148.86	&	221.44\\
NINT (default) &\cellcolor{purple2}389	&	\cellcolor{purple2}1644	&	399	&	1639	&	\cellcolor{purple1}1172	&	\cellcolor{purple1}2082	&	\cellcolor{purple2}25.09	&	\cellcolor{purple1}102.88	&	\cellcolor{purple1}26.00	&	\cellcolor{purple1}102.56	&	\cellcolor{purple1}73.51	&	\cellcolor{purple1}130.21\\

\midrule
$B=40~\%$  &391	&	\cellcolor{purple1}1636	&	\cellcolor{purple1}375	&	\cellcolor{purple1}1555	&	1319	&	2245	&	28.83	&	150.84	&	27.74	&	\cellcolor{purple2}109.44	&	93.10	&	158.04\\
$B=60~\%$   &451	&	1815	&	\cellcolor{purple2}387	&	\cellcolor{purple2}1633	&	1392	&	2177	&	35.85	&	140.64	&	30.98	&	126.79	&	108.14	&	168.48\\
$B=80~\%$ &484	&	1898	&	421	&	1709	&	1461	&	2308	&	41.49	&	159.05	&	36.34	&	143.30	&	122.60	&	192.89\\

\midrule
\textit{dense2} &708	&	3110	&	608	&	2778	&	1672	&	3478	&	38.78	&	174.44	&	34.73	&	155.76	&	93.95	&	194.84\\
\textit{Incremental}  &\cellcolor{purple1}380	&	4550	&	786	&	3360	&	\cellcolor{purple2}1208	&	4450	&	\cellcolor{purple1}24.95	&	153.56	&	\cellcolor{purple2}26.32	&	183.57	&	\cellcolor{purple2}76.12	&	236.95\\
\textit{Decremental}&600	&	3082	&	404	&	3080	&	2160	&	3329	&	30.27	&	154.90	&	28.25	&	153.51	&	149.13	&	189.60\\

\midrule
\textit{Step}&\cellcolor{purple1}380	&	1719	&	404	&	1644	&	1332	&	\cellcolor{purple2}2123	&	25.21	&	\cellcolor{purple2}115.62	&	26.63	&	110.26	&	88.19	&	182.28\\
\textit{Linear}&394	&	1737	&	398	&	1641	&	1302	&	2185	&	26.32	&	118.79	&	26.59	&	111.84	&	87.32	&	\cellcolor{purple2}151.98\\

\bottomrule
    \end{tabular}
    \caption
    {Ablation study on the impact of batch size $B$, sampling interval, and batch size scheduler in NINT, showing iterations and runtime (in seconds) required to reach target thresholds for PSNR, SSIM, and LPIPS on image fitting tasks.
    \colorbox{purple1}{Purple}: the best performance; \colorbox{purple2}{Pale Purple}: the secondary performance.}
    \label{table:settings}
\end{table*}

In this section, we provide extended experimental results on 2D image fitting tasks.
Fig. \ref{fig:div2k} shows a visualized comparison with Stand., Unif., EVOS~\cite{zhang2025evos}, and Expan.~\cite{zhang2025expansive} on training multiple images from DIV2K~\cite{agustsson2017ntire} dataset for fixed 60 seconds. Across all images, NINT consistently achieves the best reconstruction quality, which is also corroborated by the final PSNR. This confirms the indispensability of incorporating NTK-guided sample selection into INR tasks to establish a \textit{state-of-the-art} paradigm.

\begin{figure}[t]
\centering
\includegraphics[width=0.48\textwidth]{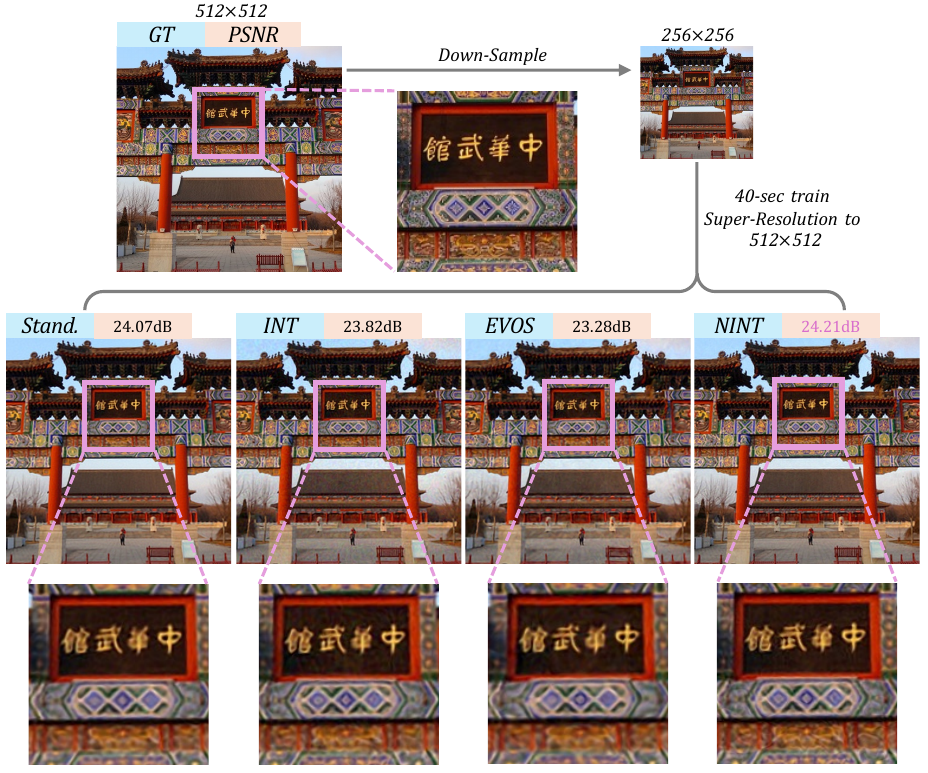}
\caption{Visual comparison of \textbf{super-resolution} results.}
\label{fig:chungwah}
\end{figure}

We also follow INT \cite{zhang2024nonparametric} to provide ablation studies on NINT settings including batch size $B$, sampling interval, and batch size scheduler. 
By default, NINT adopts: \textbf{(1)}~batch size of $B\,=20\,\%$, hence we provide comparisons of $B\,=40\,\%$, $B\,=60\,\%$, and $B\,=80\,\%$; \textbf{(2)}~sampling interval of \textit{dense} (sample at every iteration), hence we provide comparisons of \textit{dense2} (sample at every other iteration), \textit{Incremental}, and \textit{Decremental}; \textbf{(3)}~constant batch size, hence we provide comparisons of \textit{Step} batch size scheduler and \textit{Linear} batch size scheduler.

As Table \ref{table:settings} shows, the default NINT configuration consistently achieves the fastest convergence in most cases, attaining the best or second-best performance in 9 out of 12 metrics. 
Larger batch sizes can barely improve performances in most cases, suggesting NINT's superiority of exploiting NTK's effectiveness for training acceleration.
Among interval settings, \textit{Incremental} (gradually increasing sampling frequency) yields the fastest PSNR\,$=\,30$ convergence (380 iterations), while the default \textit{dense} remains most balanced overall. 
For batch size schedulers, the simple \textit{Step} variant demonstrates slight improvements, yet the default constant batch size remains clearly competitive.
Overall, these ablations confirm that NINT’s core design, NTK-guided selection, is the primary driver of acceleration, with auxiliary choices playing secondary roles. 

\section{Super-resolution Task}

We also demonstrate the effectiveness of NINT on \textbf{super-resolution} experiments on DIV2K dataset: downsample 512$\times$512 $\to$ 256$\times$256, fit INR only on low-resolution input, then evaluate high-resolution  reconstruction on original ground truth. This task inherently requires the model to recover missing high-frequency details and implicitly ``inpaint'' structured information.
As shown in Table~\ref{table:super} and Figure~\ref{fig:chungwah}, NINT outperforms Stand. / INT / EVOS sampling in final PSNR/SSIM/LPIPS and reaches target quality (e.g. 20 dB PSNR) much faster.

\begin{table}[h]
    \centering
    {\small
    \resizebox{0.95\linewidth}{!}{%
    \begin{tabular}{l|ccc|c}
        \toprule
        & \multicolumn{3}{c|}{1000 Iterations} & \multirow{2}{*}{ Time (s) $\downarrow$}\\ 
        Strategy & PSNR $\uparrow$ &SSIM $\uparrow$ &LPIPS $\downarrow$ & \\
        \midrule
        \rowcolor{gray2}
        Stand. & 24.390 & 0.797 & 0.301 & 8.63 \\
        INT \cite{zhang2024nonparametric}  & 24.196 & 0.760 & 0.335 & 10.82 \\
        EVOS \cite{zhang2025evos} & 23.645 & 0.740 & 0.379 & 11.56 \\
        NINT   & \cellcolor{purple1}24.266 & \cellcolor{purple1}0.782 & \cellcolor{purple1}0.310 & \cellcolor{purple1}7.91 \\
        \bottomrule
    \end{tabular}%
    }}
    \caption{\textbf{Super-resolution} performance at 1000 iterations and time to PSNR = 20 dB. \colorbox{purple1}{Purple}: the best performance.}
    \label{table:super}
\end{table}

\section{Large-scale Image Fitting Task}
\begin{figure}[h]
\centering
\includegraphics[width=\linewidth]{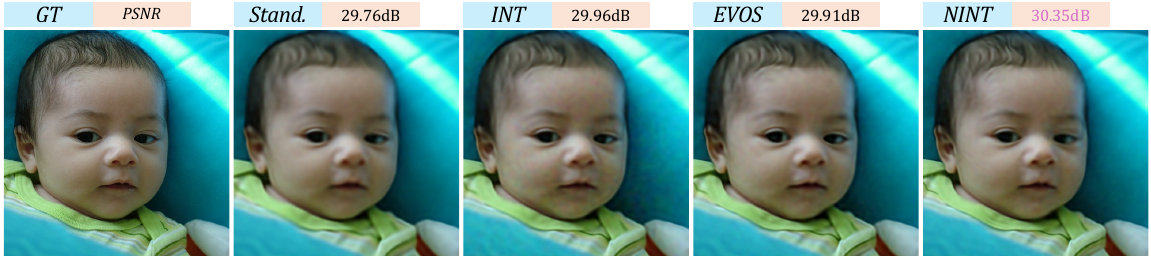}
\caption{Visual comparison of \textbf{large-scale}  image fitting results.}
\label{fig:baby}
\end{figure}

We test NINT on 1024$\times$1024 image fitting experiments on FFHQ dataset (Table~\ref{table:mega} and Figure~\ref{fig:baby}). NINT reaches 30\,dB PSNR fastest ($\sim$31\% time reduction vs. Stand.) with visibly sharper details. The core heterogeneous structure of the NTK (varying self-leverage and coupling) remains consistent across scales, no large-image-specific patterns were observed.

\begin{table}[h]
\centering
    {\small
    \resizebox{0.95\linewidth}{!}{%
    \begin{tabular}{l|ccc|c}
        \toprule
                & \multicolumn{3}{c|}{1000 Iterations} & \multirow{2}{*}{ Time (s) $\downarrow$}\\ 
        Strategy & PSNR $\uparrow$ &SSIM $\uparrow$ &LPIPS $\downarrow$ & \\
        \midrule
        \rowcolor{gray2}
        Stand. & 31.702 & 0.808	& 0.413	& 141.53 \\
        \midrule
        INT \cite{zhang2024nonparametric}& 31.767	&0.798	&0.417& 123.19\\
        EVOS \cite{zhang2025evos} & 31.903	&0.810	&0.404	& 124.33\\
        NINT	&\cellcolor{purple1}31.957&\cellcolor{purple1}0.815	&\cellcolor{purple1}0.396	&\cellcolor{purple1}  97.32  \\ 
        \bottomrule
    \end{tabular}
    }
    }
    \caption{\textbf{Large-scale} image fitting at 1000 iterations and time to PSNR~$=30$~dB. \colorbox{purple1}{Purple}: the best performance.}
     \label{table:mega}
\end{table}

\section{Efficient NTK Computation}

A potential bottleneck in NINT is the explicit construction of the Neural Tangent Kernel (NTK) matrix $K_{\theta^t} \in \mathbb{R}^{N \times N}$. As defined in Eq.~\ref{eq:ntk}, calculating the full matrix is computationally heavy for high-resolution signals. 

To maintain a lightweight sampling overhead, we avoid forming $K_{\theta^t}$ explicitly. Since the NTK-guided selection of $\mathcal{B}^\star$ only involves the product of the NTK and the loss gradient vector $\mathbf{g}^t$, we implement this efficiently using two automatic differentiation primitives: a Vector-Jacobian Product (VJP) followed by a Jacobian-Vector Product (JVP). Specifically, we first compute $\mathbf{v} = \mathbf{J}^\top \mathbf{g}^t$, where $\mathbf{J}$ is the Jacobian of the model outputs w.r.t.\ parameters, and subsequently compute the scores $\mathbf{w} = \mathbf{J}\mathbf{v} = \mathbf{J}(\mathbf{J}^\top \mathbf{g}^t) = K_{\theta^t} \mathbf{g}^t$. 

In practice, it is found that an NTK scoring typically takes $3.2$ ms wall-clock time ($3.6\%$ of total selection) on an NVIDIA RTX 4090, with Kodak dataset (512$\times$512 pixels) and 5$\times$256 SIREN. 
This underscores NINT's capacity to leverage negligible extra computation into significant INR training acceleration.

\end{document}